\documentclass[letterpaper, 10 pt, conference]{ieeeconf}  

\IEEEoverridecommandlockouts                              

\usepackage{graphics} 
\usepackage{epsfig} 
\usepackage{mathptmx} 
\usepackage{times} 
\usepackage{amsmath} 
\usepackage{amssymb}  
\usepackage{bm}
\usepackage{algorithm}
\usepackage{multirow}
\usepackage{verbatim}
\usepackage{graphicx}
\usepackage{hyperref}
\usepackage{textcomp}
\usepackage{cite}
\usepackage{subcaption}
\usepackage{lipsum}
\usepackage{mwe}
\usepackage[font=footnotesize]{caption}

\usepackage{titlesec}
\titlespacing{\section}{0pt}{*0.6}{*0.6}
\titlespacing{\subsection}{4pt}{*0.5}{*0.5}

\IEEEoverridecommandlockouts                   
\usepackage{booktabs} 
\title{\LARGE \bf
ARviz -- An Augmented Reality-enabled Visualization Platform for ROS Applications
}

\author{Khoa C. Hoang, Wesley P. Chan, Steven Lay, Akansel Cosgun, Elizabeth A. Croft \\ Department of Electrical and Computer Systems Engineering, Monash University} 
\begin{document}

\maketitle
\thispagestyle{empty}
\pagestyle{empty}

\begin{abstract}
Current robot interfaces such as teach pendants and 2D screen displays used for task visualization and interaction often seem unintuitive and limited in terms of information flow. This compromises task efficiency as interacting with the interface can distract the user from the task at hand. Augmented Reality (AR) technology offers the capability to create visually rich displays and intuitive interaction elements in situ. In recent years, AR has shown promising potential to enable effective human-robot interaction. We introduce ARviz - a versatile, extendable AR visualization platform built for robot applications developed with the widely used Robot Operating System (ROS) framework. ARviz aims to provide both a universal visualization platform with the capability of displaying any ROS message data type in AR, as well as a multimodal user interface for interacting with robots over ROS. ARviz is built as a platform incorporating a collection of plugins that provide visualization and/or interaction components. Users can also extend the platform by implementing new plugins to suit their needs. We present three use cases as well as two potential use cases to showcase the capabilities and benefits of the ARviz platform for human-robot interaction applications. The open access source code for our ARviz platform is available at: https://github.com/hri-group/arviz.

\end{abstract}

\section{BACKGROUND AND MOTIVATION}
        \label{sec:introduction}
        
The Robot Operating System (ROS) \cite{ros} has risen to become the framework of choice for robotics research and development. ROS is used in many robotic research areas, ranging from mobile robots and robotic manipulators to aerial robotics, as well as applications ranging from factory and healthcare robots, to disaster response and space exploration robots. The industrial-compatible version ROS2 is under active development. 

\begin{figure}
    \centering
    \includegraphics[width=0.98\linewidth]{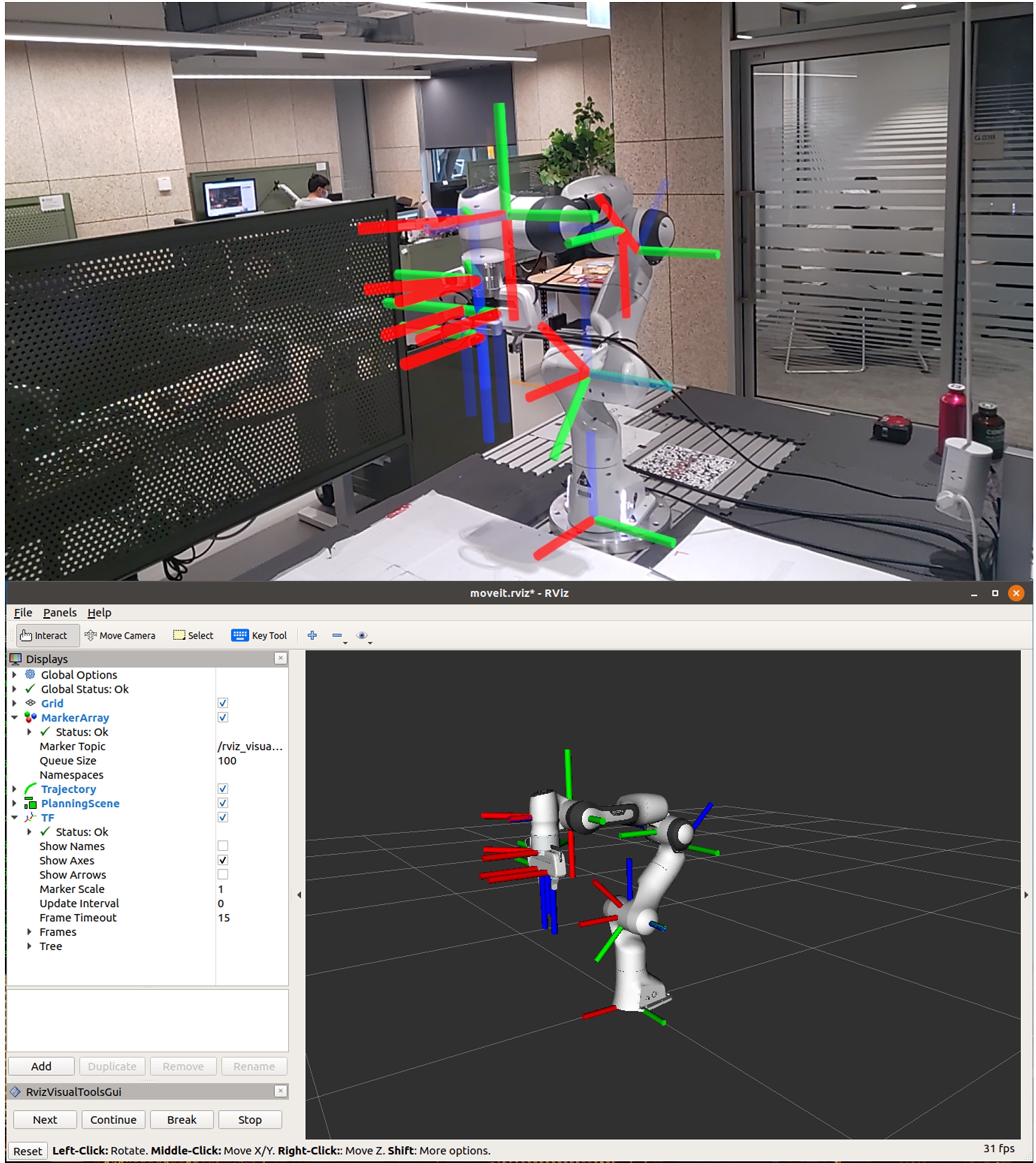}
    
    \caption{\textbf{Display Plugin} used for visualizing \textit{TF} ROS message data type on our platform ARviz (top) and on Rviz in ROS (bottom). The similarities between the two visualization tools makes using ARviz more convenient}
    \label{fig:rvizarviz}
\end{figure}

When working with robots, the ability for users to visualize sensor data, understand robot state, and issue commands in an intuitive way is crucial. The successful software tool, RViz\footnote{rviz - http://wiki.ros.org/rviz}, offers these capabilities for the development of robotics applications built on the ROS framework through a 2D, screen-based visual interface (Figure \ref{fig:rvizarviz} (bottom)). However, providing visualizations and interfaces in a 2D screen for a 3D world has inevitable challenges with respect to intuitiveness (especially with respect to spatial understanding) for the user. Furthermore, in order to see the visualization/information or to command the robot, a screen-based interface requires the user to frequently shift their attention to the monitor screen, disconnecting them from the physical robot/environment. This attention shifting results in reduced levels of safety and efficiency when operating or collaborating with robots. 

Augmented Reality (AR) technology offers the capacity to create and superimpose virtual objects and information directly onto the real world, as viewed by the operator, co-located in the robot's space and task space. In recent years, AR technology has risen in popularity becoming a promising new alternative for creating interfaces that can facilitate more intuitive and effective interaction and information exchange in human-robot collaborations \cite{Green}. AR-based interfaces offer a richer visual to the user and more intuitive interaction through virtual objects superimposed onto the real world. By rendering relevant information in the user's field of view directly over the real world, AR interfaces help reduce mental fatigue and cognitive load of the user when collaborating with their robot partner \cite{brending}. 

Recognizing the potential of AR technology, researchers have proposed various AR-based interfaces to assist human robot interaction over the past few years. Yew et al.\cite{yew} developed an AR-based teleoperation interface for a maintenance robot, where a mixture of real and virtual objects were used to reconstruct the remote maintenance site at the operator's environment. Makhataeva et al.\cite{makhataeva} increased a human operator's awareness of danger in human-robot collaboration by augmenting their views through an AR display with a safety aura around the operating robot. Walker et al.\cite{walker} used an AR display to show various visual cues such as arrows and points to convey motion intent of aerial robots during collaboration between human and robot. Dinh et al.\cite{dinh} proposed a novel interface for collaborating with a tape dispensing robot. Their system allowed the operator to monitor the task through an AR headset, and enhanced the user's experience working with the robot. Waymouth et al.\cite{waymouth2021demonstrating} developed an interaction design for demonstrating cloth folding to robots through AR. Chan et al.\cite{chan} utilized AR in conjunction with gestures and tactile feedback to create a multimodal system for robot programming and execution, where the AR display was used to assist the human operator with robot programming in addition to visualizing robot trajectories. A similar system was later on applied towards large-scale human-robot collaborative manufacturing tasks \cite{chan2020}.

The number of AR-HRI research papers in recent years increased rapidly from only 4000 papers in 2015 to 8000 papers in 2019\cite{ARReview}. However, most of these AR applications developed are task-specific (including works mentioned above), lacking generalizability, reusability, and scalability; i.e., the AR-HRI applications developed for these tasks are not easily re-purposed for other tasks.  To enable more efficient application development and research on AR-enabled robotics and HRI, a universal platform that enables a systematic approach to implementing data visualization and interactive interfaces for AR-HRI applications is proposed to fill an unmet need for the AR-HRI community.

Recognizing this need, we developed ARviz -- a versatile AR visualization platform for ROS-based applications. ARviz is inspired by Rviz, and allows users to visualize any standard ROS messages data types in AR. Furthermore, with additional input modalities, such as speech, gesture, and gaze, enabled by AR devices, ARviz provides a multimodal interface for users to interact with any ROS applications. ARviz is designed to be extendable, allowing users to implement their own \textit{plugins} for visualizing customized ROS messages, or for defining their own interaction methods based on their specific application needs.
By keeping many design similarities with Rviz, ARviz aims to provide a sense of familiarity for the large existing user-base of Rviz, as shown in Figure \ref{fig:rvizarviz}. 
In the following, Section \ref{sec:system_design} describes ARviz's platform architecture. Section \ref{sec:available_plugins} presents the collection of display and tool plugins included in our current implementation of ARviz. Section \ref{sec:use_cases} demonstrates the benefits of ARviz through a number of use cases featuring different applications of ARviz. Finally, Section \ref{sec:discussion} discusses potential further extensions and uses cases for ARviz, as well as some limitations, while Section \ref{sec:conclusion} summarises the ARviz contribution. 
\section{PLATFORM ARCHITECTURE}
        \label{sec:system_design}
        \begin{figure*}[h]
    \centering
    \includegraphics[width=0.98\linewidth]{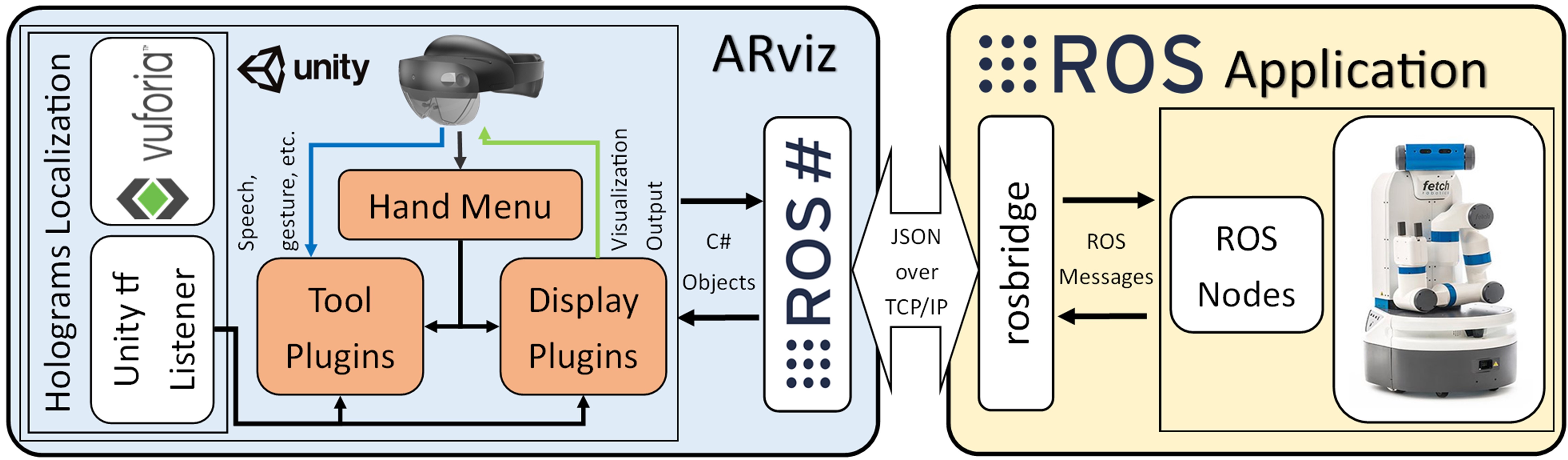}
    
    \caption{Basic Architecture of ARviz Platform}
    \label{fig:sysDesign}
\end{figure*}

ARviz is designed to be a stand-alone AR application that can connect to any ROS-based application to provide visualization and interaction capabilities to the user, requiring minimal-to-no-modifications to the existing ROS application. Figure \ref{fig:sysDesign} depicts the overall platform architecture for ARViz and how it interacts with ROS applications. 

We developed ARViz using the Unity Game Engine. ARviz is designed to work with many different plugins, which can be categorized into two types: \textbf{Display Plugins}, which are used for visualizing data received from ROS applications, and \textbf{Tool Plugins}, which enable the use of  natural input methods such as speech and gestures for interacting with ROS applications. The visual elements (holograms) generated by the plugins are accurately localized in the real world with the help of \textit{Vuforia Engine}\footnote{Vuforia Engine: https://www.ptc.com/en/products/vuforia/vuforia-engine} - a Unity toolkit that enables the alignment of the virtual world and the real world, and \textbf{Unity tf Listener} - a component of ARviz that collects spatial data (transformations between coordinate frames) from the ROS application. ARviz provides a \textbf{Hand Menu}, which can be accessed with a raise of hand, as a convenient means to manage the plugins available on the platform (Figure \ref{fig:hand_menu}). The communication channel between ARviz and the ROS application is established using the Unity package \textit{ROS\#}\footnote{Siemens - ROS\#: https://github.com/siemens/ros-sharp}on the ARViz side and a ROS package \textit{rosbridge}\footnote{rosbridge\_suite: http://wiki.ros.org/rosbridge\_suite} running on top of the ROS application. We explain each component in more detail in the following subsections.
        \subsection{ARviz Plugins}
            \label{subsec:plugins}  
            \textbf{Display Plugins} are components of the ARviz platform that allow users to visualize data from ROS applications in AR by rendering virtual displays that are contextually located in the real world. \textbf{Display Plugins} can be used for visualizing application information such as robot transformation frames, navigation paths, and robot gripper grasp poses (more details in Section \ref{subsec:UC1}, \ref{subsec:UC2} and \ref{subsec:UC3}). The plugins are designed to be reusable for any ROS application using the same standard ROS message types, and users can also implement additional customized plugins to visualize other ROS message types or customized ROS message types.


Data from ROS is collected by \textit{ROS\#} through a component call a subscriber (more details in Section \ref{subsec:ros_communication}). The \textbf{Display Plugin} accesses the collected data from the subscriber and renders corresponding visual elements at 20 Hz. This callback frequency was chosen to suit the limited processing capabilities of AR headsets, whilst being frequent enough to maintain a smooth motion of the visual elements - only slightly slower than the 24 Hz frame rate for movies.

\textbf{Tool Plugins} are components of ARviz that utilize input modalities of the AR device such as gaze tracking and speech recognition to provide a multimodal communication interface with ROS. \textbf{Tool Plugins} can be used for communicating with a robot's navigation stack to move the robot around the environment or for issuing command to a robotic manipulator to start a handover task (more details are provided in Section \ref{subsec:UC1} and \ref{subsec:UC3}). Similar to \textbf{Display Plugins}, users can implement their own plugins to tailor to their needs for communicating with ROS and interacting with the robot.

The implementation of this plugin can be varied depending on the communication channel. The \textbf{Tool Plugin} is developed as an extension of the input handler provided by Microsoft's \textit{Mixed Reality Toolkit}\footnote{Microsoft Mixed Reality Toolkit Unity - https://docs.microsoft.com/en-gb/windows/mixed-reality/mrtk-unity/?view=mrtkunity-2021-05} - a Unity package for mixed reality application development which provides APIs to input system and other building blocks for spatial interactions. User inputs are collected by \textbf{Tool Plugins}, which then send the information to ROS using \textit{ROS\#}, through a component called a publisher (more details in Section \ref{subsec:ros_communication}). 
        \subsection{Hand Menu}
            \label{subsec:handmenu}           
\begin{figure}
    \centering
    \includegraphics[trim= 200 0 200 100,clip, width=0.98\linewidth]{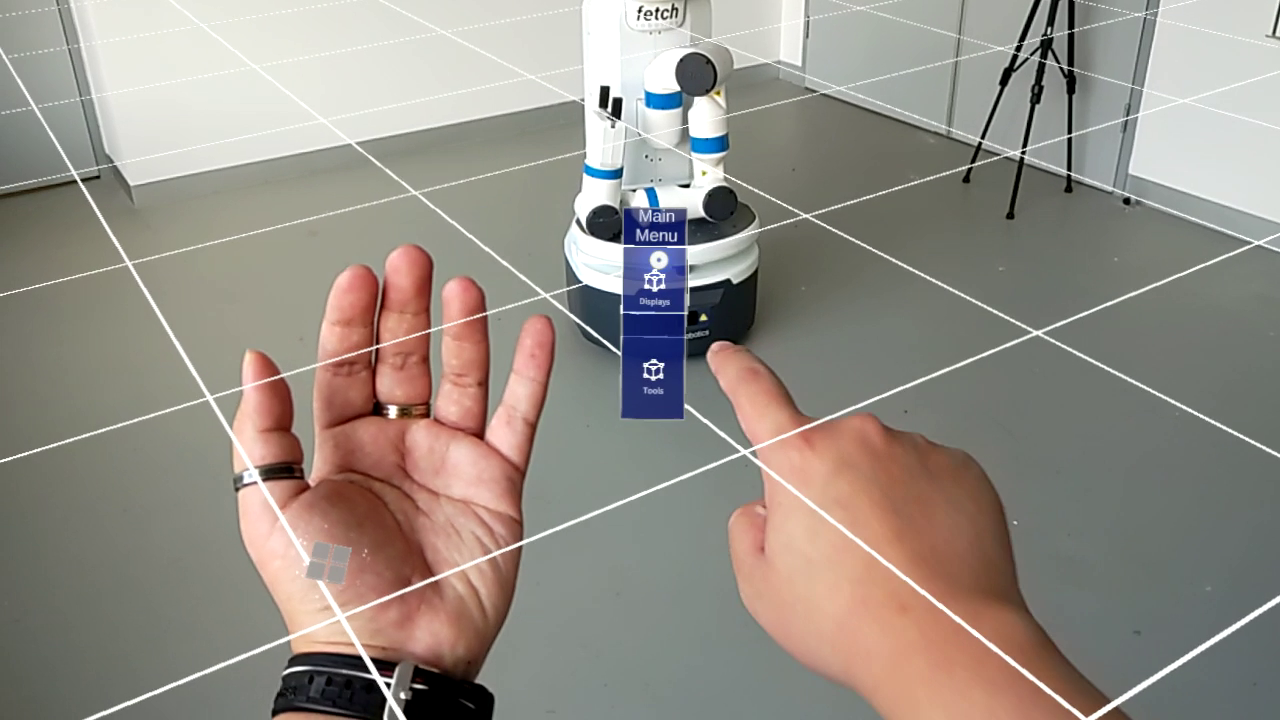}
    
    \caption{\textbf{Hand Menu} appears when the user raise their left hand in front of the HoloLens 2. \textbf{Hand Menu} provides quick access to the plugins available on the platform.}
    \label{fig:hand_menu}
\end{figure}

            A menu that can be summoned at the user's fingertips, literally, is developed to provide quick and easy control of all the available plugins implemented in ARviz. By raising their left hand as shown in Figure \ref{fig:hand_menu}, the user can make the hand menu appear and access configurations of the \textbf{Display Plugins} and \textbf{Tool Plugins}. Using this menu, users can toggle the visibility of \textbf{Display Plugins}, enable/disable \textbf{Tool Plugins} or access/modify each plugin's specific properties, as shown in Figure \ref{fig:tfDisplay}. Additional menu entries can be added and their behaviors customized for user-defined display plugins and tool plugins.

        \subsection{Hologram Localization}
            \label{subsec:holograms_localization}
            For AR applications, we need to consider two coordinate systems: the \textbf{virtual world coordinate system (VWCS)}, where virtual objects/holograms are rendered, and the \textbf{real world coordinate system (RWCS)}, where the physical entities (e.g., the robot) exist. Typically, when an AR application is started up, it arbitrarily initializes the base frame of the VWCS to be where the AR device is physically located in the real world. On the other hand, in a ROS application, the RWCS is managed by the \textit{tf} transformation ROS package, and the base frame of the RWCS can be defined by the user to be anywhere in the physical world. In order for the \textbf{Display Plugins} and \textbf{Tool Plugins} in ARviz to render virtual display elements at appropriate locations in the real world, two pieces of information are needed: 1) The pose (position and orientation) of the virtual world coordinate system (VWCS) with respect to the real world coordinate system (RWCS) - i.e., the transformation between VWCS and RWCS. 2) Where should the display element be located with respect to the real world coordinate system. These two pieces of information are provided by \textit{Vuforia Engine} and \textbf{Unity tf Listener} as described below.

\begin{figure*}
    \centering
    \includegraphics[width=0.98\linewidth]{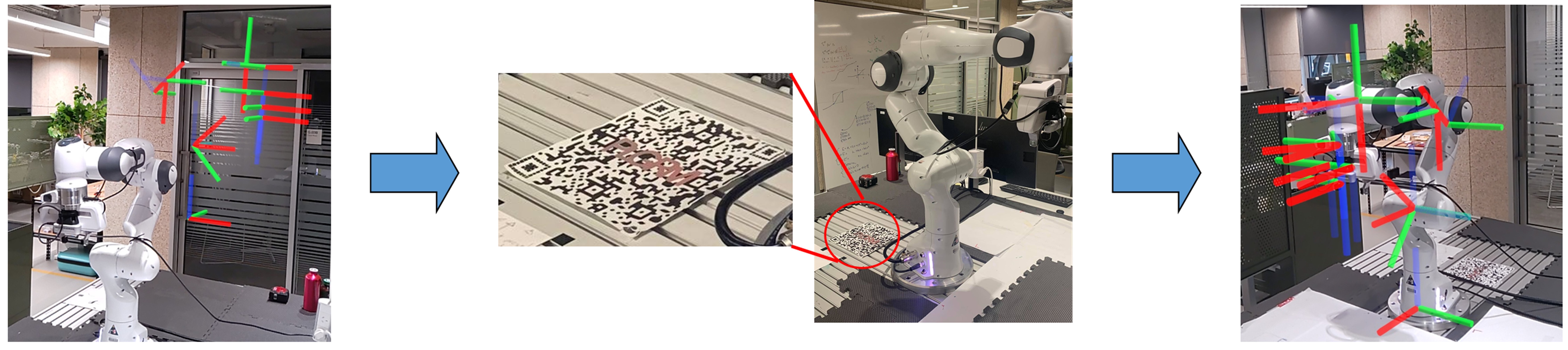}
    
    \caption{ARviz's Virtual World Alignment. The virtual objects would be initially rendered at random location, misaligning with the real world (left). With the use of the QR code marker in the real world and \textit{Vuforia Engine} in ARviz (middle), the virtual world is aligned with the real world, allowing the correct overlay of the virtual objects onto the real world.}
    \label{fig:localization}
\end{figure*}

\textbf{Aligning Real and Virtual World Coordinate Systems using Vuforia}: 
To define the transformation between RWCS and VWCS, we define an anchor point that tell us where the VWCS base frame should be, within the RWCS. \textit{Vuforia Engine} allows the specification of this anchor through a physical QR code marker placed in the real world.
To create this anchor, a physical QR code marker is placed at fixed, known location within the RWCS. In ARviz, the \textit{Vuforia Engine} is deployed to detect this QR code marker. Once the AR devices detects the QR marker, it computes the transformation between the AR device and the physical QR marker. Since the position of the AR device with respect to the VWCS is always tracked by the AR application, the transformation between the VWCS and RWCS can be computed. Once this transformation is computed, ARviz then repositions the VWCS base frame to coincide with the RWCS base frame, aligning the virtual world with the real world. The alignment process is illustrated in Figure \ref{fig:localization}.

\textbf{Unity tf Listener}:  
Data in ROS are specified with respect to to a given base frame to provide physical spatial context 
(e.g., a robot's coordinate may be specified relative to the map frame).
ROS is a distributed computing system by design. Hence, in ROS applications, transformations between different coordinate frames are asynchronously broadcasted (published) by different processes (nodes). The ROS component named \textit{tf Listener} collects and assembles all these transformation information and allows any component of the ROS application to query transformation between any two coordinate frames. The \textbf{Unity tf Listener} provides the analogous functionality in ARviz. Asynchronous transformation information from the ROS application is passed on to the \textbf{Unity tf Listener} through ROS\#. The \textbf{Unity tf Listener} collects and assembles the transformation information. 
\textbf{Display Plugins} and \textbf{Tool Plugins} can then query transformation between any two coordinate frames, and render display elements at appropriate locations. 
        \subsection{Communication with ROS}
            \label{subsec:ros_communication}
            For the plugins to receive and send data to and from ROS applications, a connection with the ROS environment is required. ROS runs on Ubuntu, and within ROS applications, \textit{nodes} (i.e, processes) communicate with each other by broadcasting \textit{messages} (i.e., data structures) over \textit{topics} (i.e., communication channels). \textit{Publishers} are used to broadcast \textit{messages} and \textit{subscribers} are used to receive and manage \textit{messages}. On the other hand, ARviz is built on the Unity Game Engine, a platform which does not natively support ROS APIs for establishing the communication channels mentioned above. To resolve this problem, a combination of \textit{rosbridge} on ROS and \textit{ROS\#} on Unity was used. \textit{rosbridge} and \textit{ROS\#} allow ARviz and any ROS application to communicate via a common standard JSON string format over TCP/IP, as shown in Figure \ref{fig:sysDesign}. \textit{rosbridge} establishes the network port for ARviz to connect to. \textit{ROS\#} provides ROS-like APIs such as \textit{ROS\# Publisher} and \textit{ROS\# Subscriber}, enabling ARviz plugins to communicate with other ROS applications through ROS-like communication mechanisms.

\section{Available Plugins}
        \label{sec:available_plugins}
\begin{figure*}[h]
    \centering
    \includegraphics[width=0.98\linewidth]{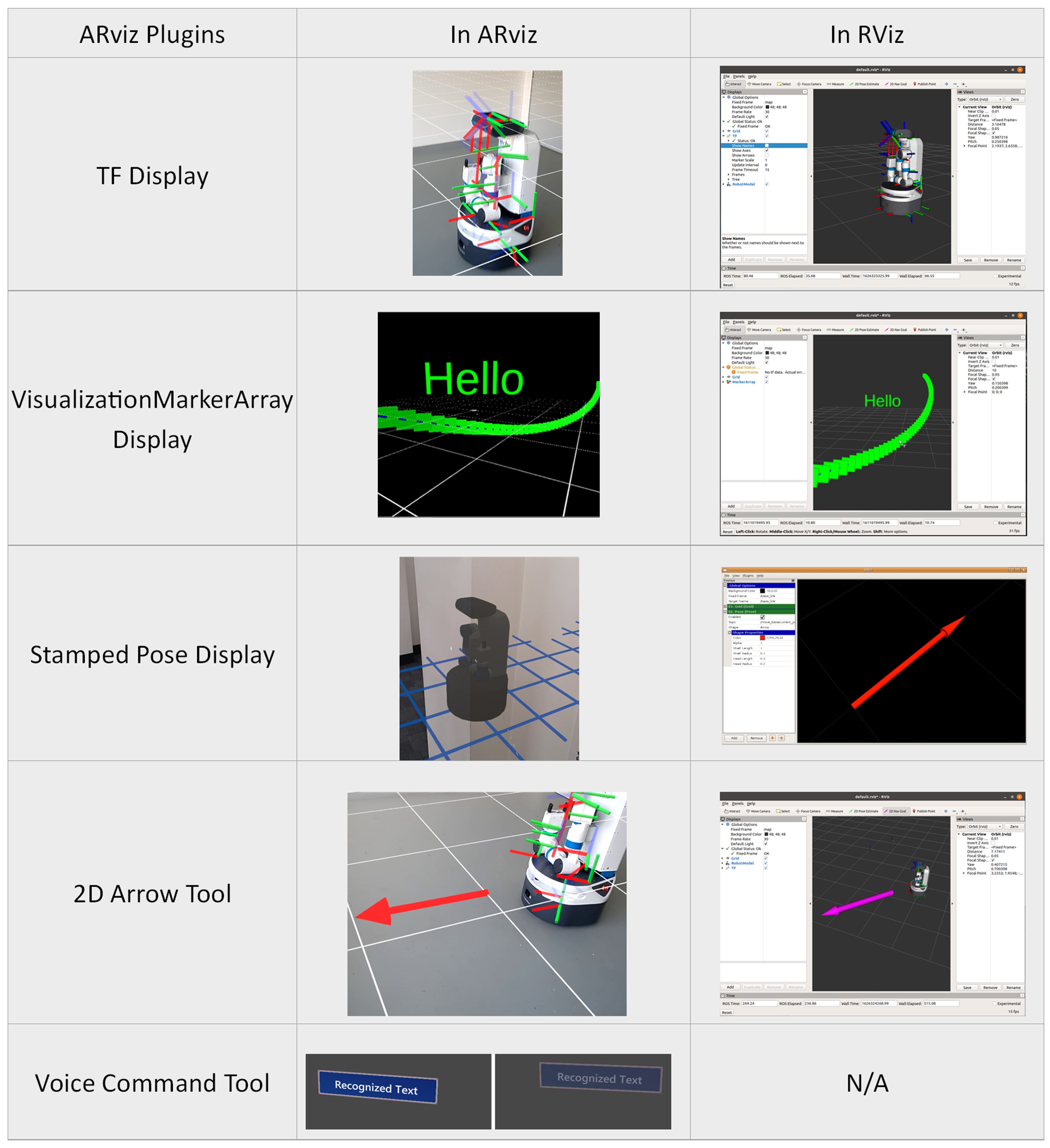}
    
    \caption{Summary of available plugins, comparing the similarity between ARviz in AR and Rviz in ROS.}
    \label{fig:pluginList}
\end{figure*}

        Three \textbf{Display Plugins} and two \textbf{Tool Plugins} were implemented in our initial release of ARviz. These plugins are developed and tested on the AR-head-mounted-display Microsoft HoloLens 2\footnote{HoloLens 2 - https://www.microsoft.com/en-AU/hololens/hardware} and are summarised in Figure \ref{fig:pluginList}.

\begin{figure}
    \centering
    \includegraphics[width=0.98\linewidth]{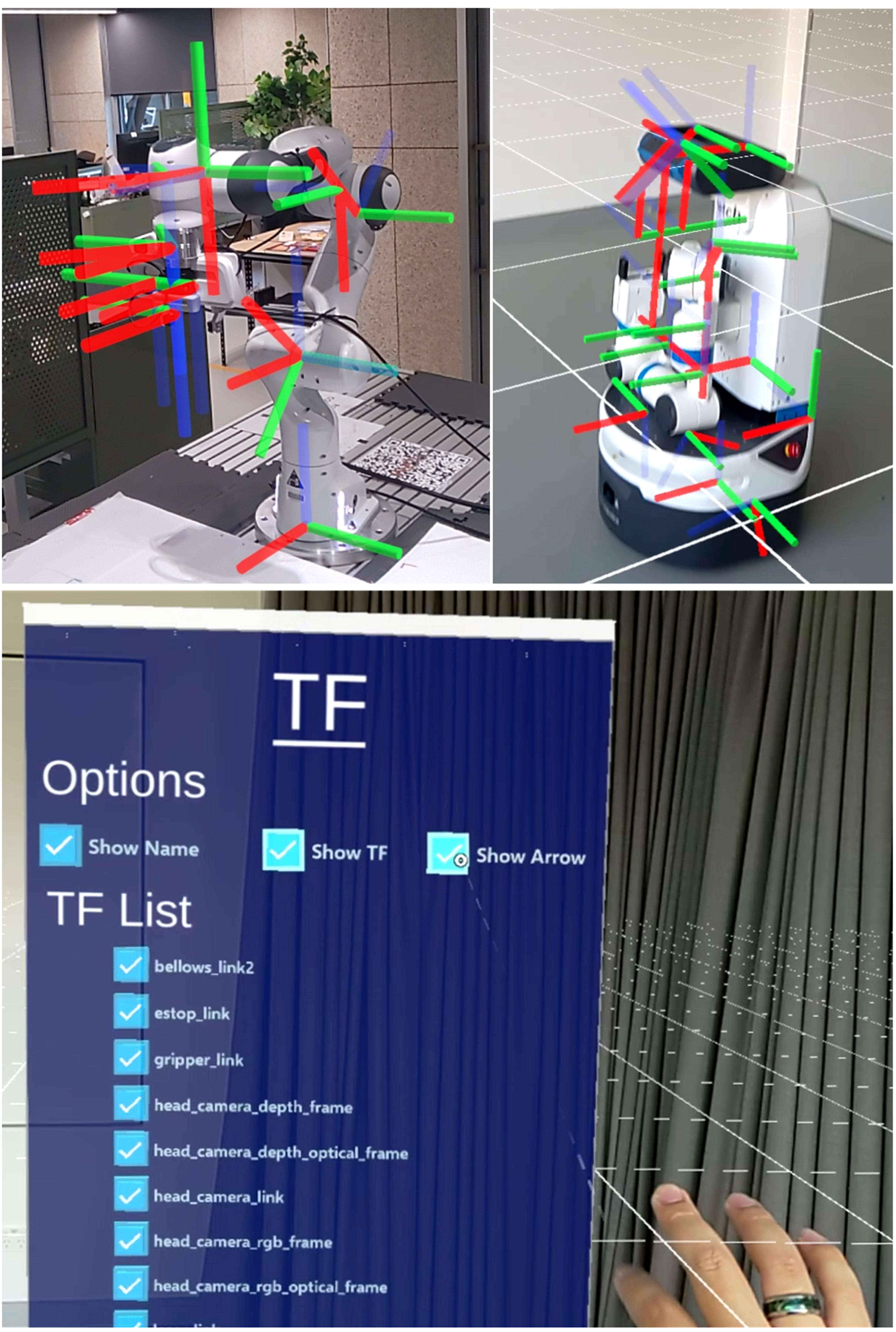}
    
    \caption{\textbf{\textit{TF} Display Plugin} in ARviz. Top: \textbf{\textit{TF} Display Plugin} being used with different robots. Bottom: \textbf{\textit{TF} Display Plugin} control panel, allows users to toggle the visibility of individual frames or all the frames at once, as well as other elements such as frame names and arrows linking the frames}
    \label{fig:tfDisplay}
\end{figure}

\textbf{\textit{TF} Display Plugin} visualizes coordinate frames monitored by the \textbf{Unity tf Listener}. Each coordinate frame is visualized as a set of 3D axes, with red, green and blue corresponding to x, y and z axis. Figure \ref{fig:tfDisplay} (top) shows examples of visualizing the joint/link frames of two robots. The plugin can also show the name of each frame and the connection between it and its parent frame, depicted with an arrow. The visibility of each elements of this display plugin can be toggled through a control panel as shown in Figure \ref{fig:tfDisplay} (bottom). \textbf{\textit{TF} Display Plugin} can be used for visually confirming the connectivity between ARviz and the ROS applicaiton, as well as verifying the alignment between the VWCS and the RWVC. Moreover, \textbf{\textit{TF} Display Plugin} allows users to visualize many aspects of robot operation, including robot localization and robot structure, and serves as a powerful development and debugging tool. Users can utilise \textbf{\textit{TF} Display Plugin} to view the connections between the coordinate frames within the robot for a better understanding of how the robot base, its multi-link manipulator and other degrees of freedom, operate as a whole. 

\textbf{\textit{VisualizationMarkerArray} Display Plugin} visualizes a collection of user-defined markers. These markers include various primitives such as cubes and spheres, text, or lines. This plugin is inspired by the \textit{MarkerArray}\footnote{rviz Markers - http://wiki.ros.org/rviz/DisplayTypes/Marker} display in Rviz, and provides the visualization capabilities similar to its rviz counterpart. \textbf{\textit{VisualizationMarkerArray} Display Plugin} allows users to display arbitrary primitives in the virtual world through a user-defined ROS \textit{topic} by publishing messages of type \textit{visualization\_msgs/MarkerArray} to this topic. This plugin provides a generic tool for creating visual markers with user-defined size, color, shape, and location. This tool can be used for a wide range of tasks such as  creating text labels, visualizing robot paths, and generating object indicators.

\textbf{\textit{Stamped Pose} Display Plugin} visualizes the translation and orientation of a virtual object with respect to a given coordinate frame, given in the format of a \textit{geometry\_msgs/PoseStamped} ROS message type. The default virtual object used for the visualization is an arrow, but users can customize this plugin by defining their own pre-loaded object model or mesh to be displayed instead the default arrow. The plugin listens for any incoming messages being published to a topic defined by the users and visualizes the pose with the virtual object accordingly. This plugin is useful for visualizing the 3D pose of robots or occluded objects for example.

\textbf{\textit{2D Arrow} Tool Plugin} allows users to specify a 2D pose (position and orientation) by placing an arrow on the ground plane in AR through the use of hand gestures. When the user raises their hand, a ray emanating from the hand is displayed in AR as a dotted line. When the user taps the index finger with the thumb (known as an ``air tap`` gesture), the location where the ray intersects the ground plane is identify, and a an arrow is placed with its tail endpoint fixed at this location.  
Next, the user can rotate the arrow on the floor by moving the raised hand - the arrow tip will follow the intersecting point between the hand ray and the ground plane. By performing another ``air tap`` gesture, the arrow is fixed and its pose is sent to be published in a user-specified ROS \textit{topic}.

\textbf{\textit{Voice Command} Tool Plugin} extends the speech handling capabilities of the Mixed Reality Toolkit and allows users to interact with the ROS application through speech commands. When the user says a pre-defined keyword which is identified by the AR device, the plugin will publish the corresponding command code into a topic specified by the user.

\section{USE CASES}
        \label{sec:use_cases}
        
In this section, we present a use case demonstration in Section \ref{subsec:UC1}. Section \ref{subsec:UC2} and Section \ref{subsec:UC3} highlight two studies utilizing ARviz, and demonstrate the usability and benefits of the platform for robotics research studies. Full details of the studies mentioned in Section \ref{subsec:UC2} and Section \ref{subsec:UC3} can be found in the cited works, while in the following subsections, we focus on how the ARviz platform enabled these studies, and summarizing the studies' key findings related to the use of AR elements provided by ARviz.

\subsection{Mobile Robot Navigation}
    \label{subsec:UC1}
    
\begin{figure}
    \centering
    \includegraphics[width=0.98\linewidth]{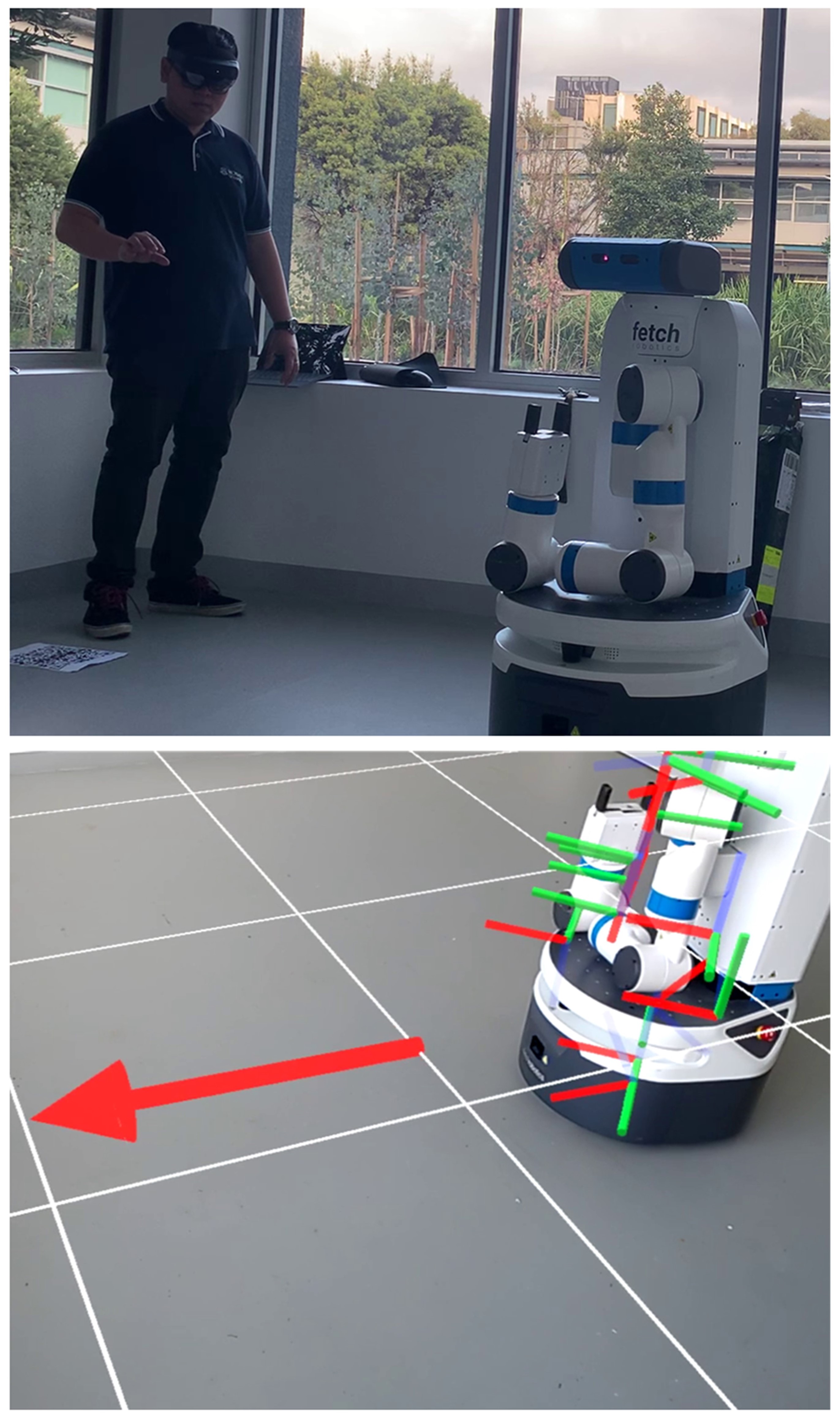}
    
    \caption{ARviz Use Case 1 - Navigating a Mobile Robot. Top: The user using ARviz with HoloLens2, navigating the \textbf{Fetch Mobile Robot}. Bottom: The augmented reality view with \textit{TF Display} overlaid on top and \textit{2D Navigation Tool} being used to navigate the robot.}
    \label{fig:UC1}
\end{figure}

We demonstrate the use of ARviz for mobile robot navigation with the \textbf{\textit{2D Arrow} Tool Plugin}, as shown in Figure \ref{fig:UC1} and a demonstration video\footnote{ARviz Demo - https://youtu.be/byUUtB5SWsM}. The robot uses the standard ROS Navigation stack, and it is launched and run independently from ARviz, without any modification needed. With the \textbf{\textit{2D Arrow} Tool Plugin}, users are able communicate the desired goal pose to the robot's navigation stack through a standard ROS topic, as demonstrated in Section \ref{sec:available_plugins}, allowing them to move the robot to any location. Compared to specifying the goal with Rviz through a 2D monitor, this \textbf{Tool Plugin} in ARviz offers the benefit of enabling better spatial and situational awareness of the goal position, as the user can see and specify the goal pose directly over the physical space. 

\subsection{Visualizing Occluded Robots and Communicating Motion Intent}
    \label{subsec:UC2}
\begin{figure}
    \centering
    \includegraphics[width=0.98\linewidth]{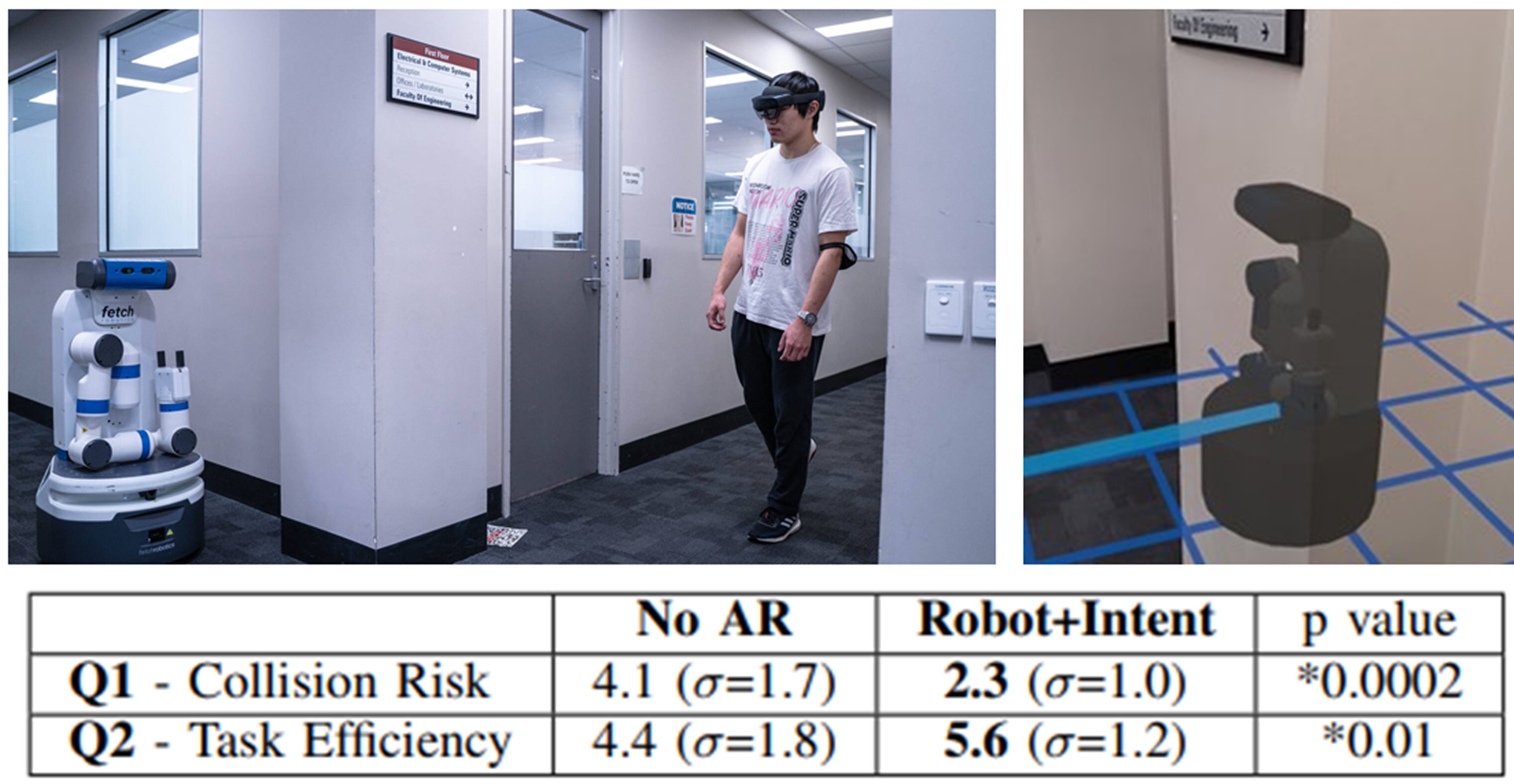}
    \caption{ARviz Use Case 2 - Visualizing Mobile Robot and Communicating its Motion Intent. Top Left: The mobile robot is not visible to the user at a T-Junction. Top Right: A representation of what the user sees from the AR headset. The robot model is visualized, enabling the user to see the robot through the wall. The cyan colored arrow emanating from the robot model showing its direction of motion. The floor is shown as a grid to improve depth perception.
    Bottom: User study survey results showing the benefit of the ARvis implementation in reducing collision risk and increasing task efficiency.(source: \cite{morris})}
    \label{fig:UC2}
\end{figure}

In the work by Gu et al. \cite{morris}, ARviz was used to visualize the location of the \textbf{Fetch Mobile Robot} using an AR headset. This allowed the user to visualize the robot even when the robot is occluded by the environment. Furthermore, the robot's direction of motion is conveyed to the user by visualizing an arrow emanating from the robot model. To achieve this, two \textbf{Display Plugins} were used to set up the AR environment, as shown in Figure \ref{fig:UC2} (top).

The first \textbf{Display Plugin} used is \textit{Stamped Pose Display}. A pre-load mesh of the mobile robot is used to customize the visualization. The second \textbf{Display Plugin} is \textit{VisualizationMarkerArray Display}. In this system, there are two types of markers used, \textit{LINE} and \textit{ARROW}. The \textit{LINE} markers are used to create the grid line on the floor, which allows better depth perception for the user wearing the AR headset. The \textit{ARROW} marker is used to visualize the motion intent of the robot. The arrow emanates from the robot model, with its tip located at the goal point the robot is moving to. The arrow is continually updated by the ROS application as the robot moves closer to the goal point.

The AR system was validated in a study with 15 participants and their experiences were surveyed with a questionnaire. The results are summarised in Figure \ref{fig:UC2} (bottom). A lower rating in perceived collision risk (2.3 with AR compared to 4.1 without AR) and higher rating in task efficiency (5.6 with AR vs 4.4 without AR) suggest that the AR system improves the experience when working in a shared workspace with the mobile robot with respect to these aspects.

\subsection{Visualizing Robotic Arm's Goal Pose During Object's Handover}
    \label{subsec:UC3}
\begin{figure}
    \centering
    \includegraphics[width=0.98\linewidth]{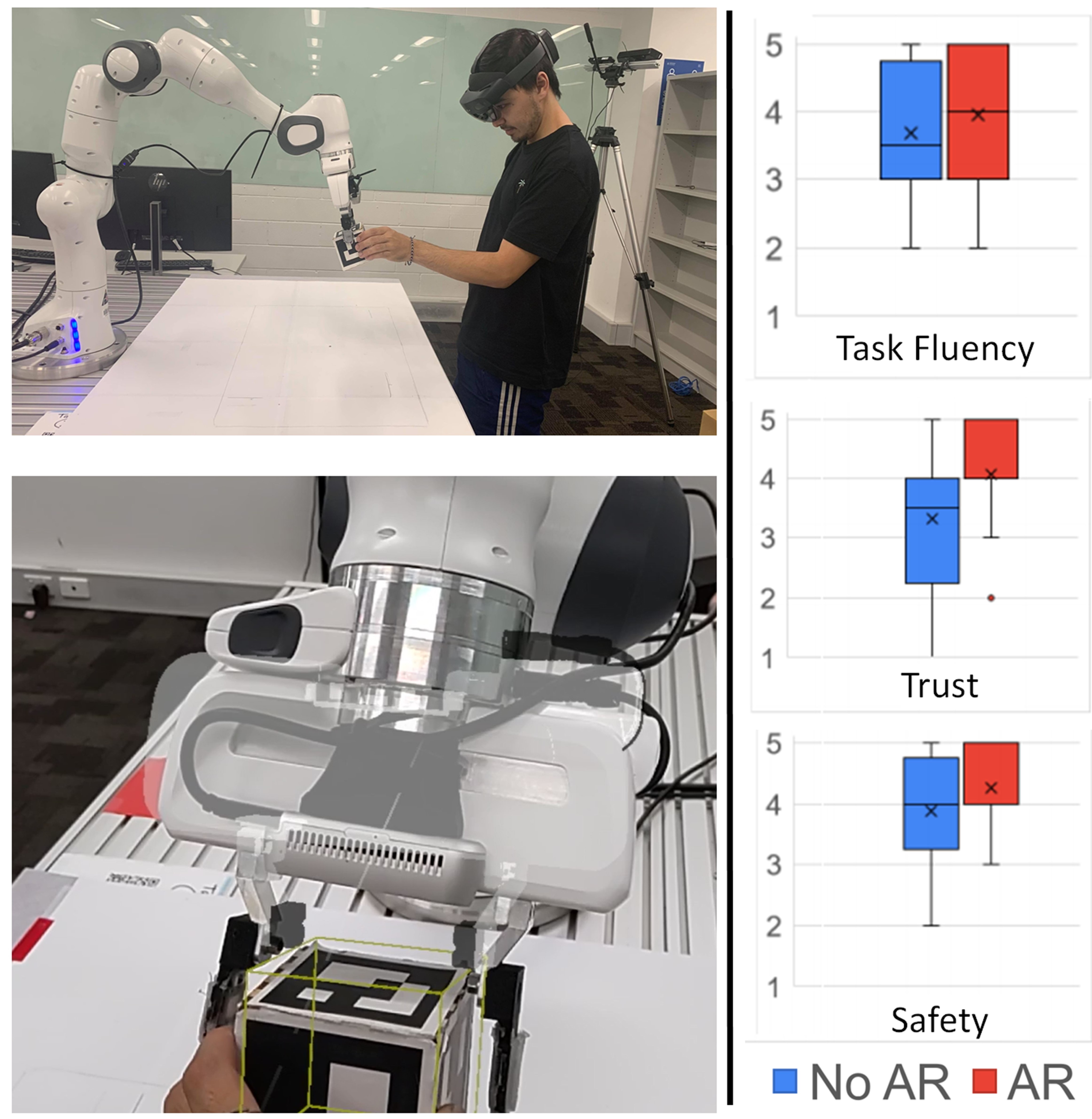}

    \caption{ARviz Use Case 3 - Visualizing a Robotic Arm's Goal Pose During Object's Handover. Top Left: The robot picking up an object from the user's hand. The user is wearing an Augmented Reality headset. Bottom Left: The detected pose of the object, and AR visualization of how the robot is planning to grasp the object. Right: Survey results from the user study reported in the paper showing that the ARvis supported interaction improved task fluency, trust and perceived saftey. (source: \cite{rhys})}
    \label{fig:UC3}
\end{figure}

In the work by Newbury et al. \cite{rhys},  ARviz was used to visualize the motion intent of the \textbf{Franka Emika Panda} robotic arm when performing handover tasks with a human collaborator. ARviz was also used to communicate to the robot when to initiate the handover task. This application (shown in Figure \ref{fig:UC3} (left)) used two \textbf{Display Plugins} and one \textbf{Tool Plugin}.

\textit{VisualizationMarkerArray Display} is used to visualize a wireframe of the object at the detected location during the handover. The wireframe is created with a collection of \textit{LINE} markers. The position and orientation of the lines are defined and updated at a fixed rate by the ROS system. The grasp pose of the robotic gripper is visualized with \textit{Stamped Pose Display}. A pre-loaded mesh of the end effector of the \textbf{Franka Emika Panda} is used to customize the \textit{Stamped Pose Display} plugin, with low opacity to avoid the virtual gripper occluding the user's view of the object during the handover in this case.

The \textbf{Tool Plugin} used by this AR application is the \textbf{\textit{Voice Command} Tool Plugin}. With this \textbf{Tool Plugin}, user can initiate the handover task by saying a pre-defined keyword. The ROS application will listen to the incoming ROS message being published by the plugin to a pre-defined topic and start the handover task as soon as the ROS message is received by the ROS application.

The AR system was validated in a study with 16 participants and their experiences were surveyed with a questionnaire given at the end of the experiment. The results were summarised in Figure \ref{fig:UC3} (right). The AR systems received a higher rating across three categories: task fluency, trust toward the robot, and perceived safety. The results suggested that the use of visualization and speech commands through ARviz both greatly improves the experience of the users during the handovers, and also leads to better efficiency in performing the task.

\section{POTENTIAL IMPLEMENTATIONS AND LIMITATIONS}
        \label{sec:discussion}
        \subsection{Potential Implementations}
    \label{subsec:potential_implementation}
Section \ref{sec:use_cases} demonstrated the capabilities and benefits of ARviz in a range of human-robot interactions such as mobile robot navigation, conveying motion intent and improving robot handover experience. Recognizing the potential of ARviz, we suggest two other robot applications that can be enabled by our platform. 

\begin{figure}
    \centering
    \includegraphics[width=0.98\linewidth]{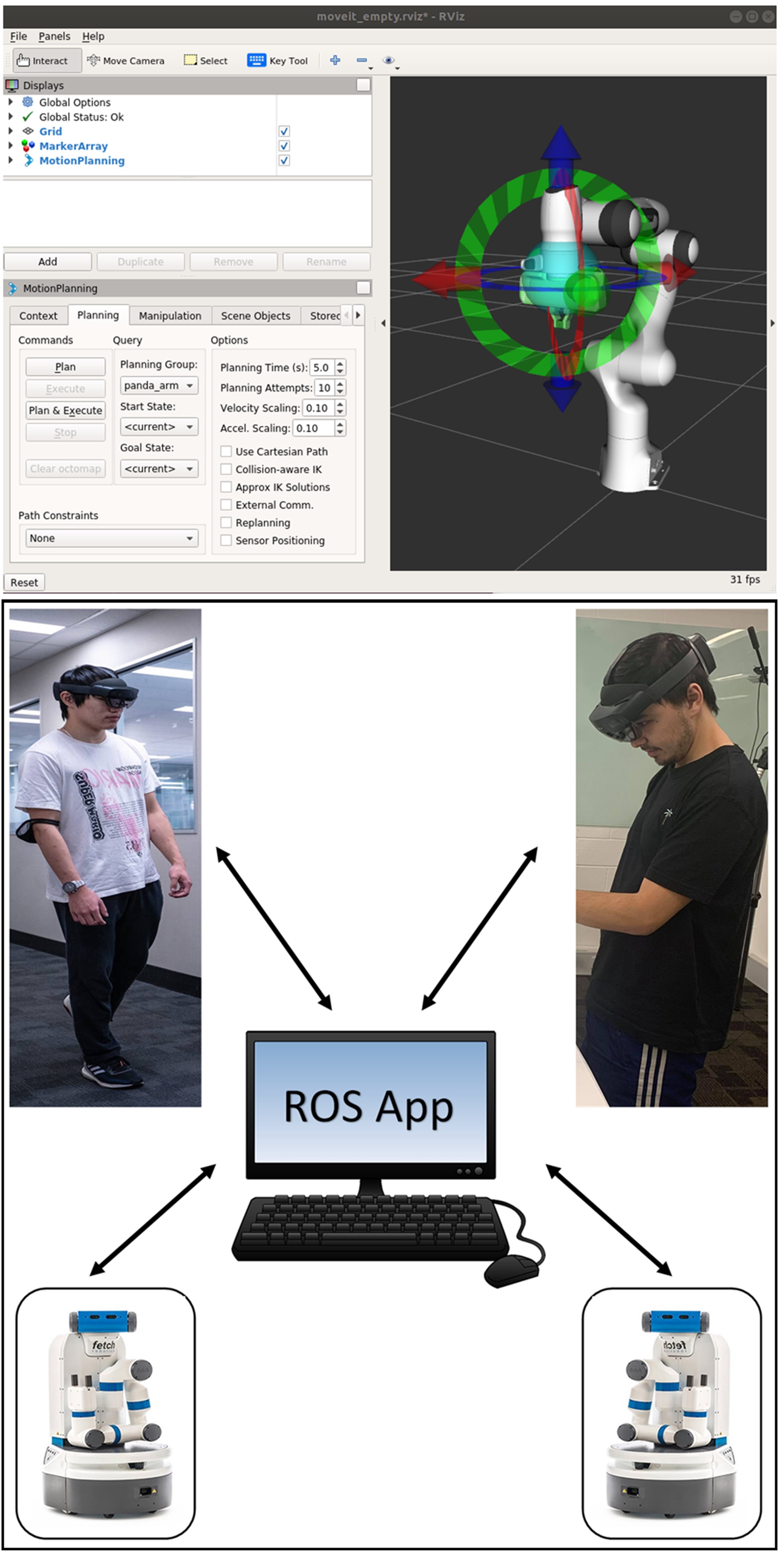}
    
    \caption{Potential Implementations with ARviz. Top: a MoveIt-inspired interactive interface with the robotic manipulator \textbf{Franka Emika Panda}. The image shows the MoveIt-implemented interface in ROS. (from: MoveIt Tutorials \cite{moveit}). Bottom: A visualization of a potential ROS application that enables multi-human-multi-robot collaborations.}
    \label{fig:potential_usecases}
\end{figure}

The first proposed application is an interactive interface for a robotic manipulator using hand gestures. In this application, ARviz could create a one-to-one virtual model of the robot, with an interactive marker that would be attached at the robot's end effector. By interacting with this marker, the operator can move the virtual robot to any desired pose (similar to how the Moveit! motion planner plugin for Rviz works, as shown in Figure \ref{fig:potential_usecases} (top)). With the virtual robot model in AR, the operator can preview directly in the robot's workspace the robot's motion, and ensure that there are no collisions with the surrounding physical environment.

The second proposed application enables multi-human-multi-robot collaboration. ARviz was designed to be a stand-alone AR software that can be deployed to any compatible AR device. That means ARviz with the same configuration settings, i.e., connecting to the same network, having the same \textbf{Display Plugins} and \textbf{Tool Plugins}, can be deployed to different AR devices and would be able to interact with the same ROS applications at the same time. An example of this would be two human operators each using ARviz on a HoloLens 2 collaborating with two mobile robots, as shown in Figure \ref{fig:potential_usecases} (bottom). ARviz is deployed on both AR devices with identical configurations, which connects to the central computer that runs the ROS applications controlling the two mobile robots. One operator can give navigation commands as described in Section \ref{subsec:UC1}. The operator can choose which robot to operate using a control panel for the navigation \textbf{Tool Plugin} or can operate both at the same time. Their collaborator (or trainee) can observe the task being implemented. The planned path of the robots can also be visualized on both AR devices, conveying motion intent of the robots to both human collaborators. 

\subsection{Limitations}
    \label{subsec:limitations}
Whilst we have had early success in demonstrating the potential of the platform in Section \ref{sec:use_cases}, ARviz still has a number of limitations requiring further development. Two key development issues are discussed here. 

The current implementation is has limited accuracy in aligning the virtual world and the real world. The semi-automatic localization method with QR-code marker provides the simplicity in implementation and usage, but the trade-off is the spatial accuracy of the overlaid visual elements, as they are affected by the accuracy of the placement of the QR code marker. An improved localization algorithm is needed in future versions of the platform.

ARviz is also limited in its ability to change the \textit{topics} that plugins connect to at run-time. The ROS \textit{topics} that \textit{ROS\#} collects data from can't be modified during run-time and must to be pre-configured prior to the deployment of the platform onto the AR device. This connectivity limitation for plugins is inconvenient for AR-based ROS applications where multiple data sources are involved - for example, a ROS application that needs to visualize data collected from multiple cameras mounted at different places on the robot. Future versions of ARviz will allow users to modify ROS \textit{topics} in run-time for one plugin, so that the visualization of multiple data sources with the same ROS message type can be achieved. A workaround for the current version would be to create multiple instances of the plugin, with each connecting to a different topic.

\section{CONCLUSION}
        \label{sec:conclusion}
        In this work, we introduced the ARviz platform, an open source platform for Augmented Reality interaction with ROS based robotic systems. ARviz supports visualization of various standardized data types that are available on ROS using \textbf{Display Plugins}. We also introduced the implementation of \textbf{Tool Plugins} that utilizes the natural input methods supported by the AR headset such as speech and gesture, to communicate with the ROS environment. ARviz provides an easy-to-access and extendable \textbf{Hand Menu} to control the plugins, allowing future developers to expand the platform. We present three different use cases of the platform from our labs, from conveying mobile robot motion intent to visualizing hand over tasks.  As demonstrated in published user studies conducted using the ARvis system, the platform provides a number of benefits, with users feeling safer when collaborating with the robot and completing the given task more efficiently. 

Finally, we proposed two different potential applications that can highlight the benefits of ARviz:  an interactive "Moveit" type interface for a robotic manipulator and an application that enables multi-human-multi-robot collaboration. Despite the success of the platform, there is still a great deal of room for improvement, including improving the accuracy of the semi-automatic virtual world alignment and addressing limitations in connecting plugins to topics at run-time. We continue to work on this exciting platform to improve capabilities and to bring a better experience to end-users.

\bibliographystyle{IEEEtran}
\bibliography{ref}

\begin{thebibliography}{10}
\providecommand{\url}[1]{#1}
\csname url@rmstyle\endcsname
\providecommand{\newblock}{\relax}
\providecommand{\bibinfo}[2]{#2}
\providecommand\BIBentrySTDinterwordspacing{\spaceskip=0pt\relax}
\providecommand\BIBentryALTinterwordstretchfactor{4}
\providecommand\BIBentryALTinterwordspacing{\spaceskip=\fontdimen2\font plus
\BIBentryALTinterwordstretchfactor\fontdimen3\font minus
  \fontdimen4\font\relax}
\providecommand\BIBforeignlanguage[2]{{%
\expandafter\ifx\csname l@#1\endcsname\relax
\typeout{** WARNING: IEEEtran.bst: No hyphenation pattern has been}%
\typeout{** loaded for the language `#1'. Using the pattern for}%
\typeout{** the default language instead.}%
\else
\language=\csname l@#1\endcsname
\fi
#2}}

\bibitem{ros}
M.~Quigley, K.~Conley, B.~Gerkey, J.~Faust, T.~Foote, J.~Leibs, R.~Wheeler,
  A.~Y. Ng, \emph{et~al.}, ``Ros: an open-source robot operating system,'' in
  \emph{ICRA workshop on open source software}, vol.~3, no. 3.2.\hskip 1em plus
  0.5em minus 0.4em\relax Kobe, Japan, 2009, p.~5.

\bibitem{Green}
S.~A. Green, M.~Billinghurst, X.~Chen, and J.~G. Chase, ``Human robot
  collaboration: An augmented reality approach—a literature review and
  analysis,'' in \emph{International Design Engineering Technical Conferences
  and Computers and Information in Engineering Conference}, 2007, pp. 117--126.

\bibitem{brending}
\BIBentryALTinterwordspacing
S.~Brending, A.~M. Khan, M.~Lawo, M.~M\"{u}ller, and P.~Zeising, ``Reducing
  anxiety while interacting with industrial robots,'' in \emph{Proceedings of
  the 2016 ACM International Symposium on Wearable Computers}, ser. ISWC
  '16.\hskip 1em plus 0.5em minus 0.4em\relax New York, NY, USA: Association
  for Computing Machinery, 2016, p. 54–55. [Online]. Available:
  \url{https://doi.org/10.1145/2971763.2971780}
\BIBentrySTDinterwordspacing

\bibitem{yew}
A.~Yew, S.~Ong, and A.~Nee, ``Immersive augmented reality environment for the
  teleoperation of maintenance robots,'' \emph{Procedia Cirp}, vol.~61, pp.
  305--310, 2017.

\bibitem{makhataeva}
Z.~Makhataeva, A.~Zhakatayev, and H.~A. Varol, ``Safety aura visualization for
  variable impedance actuated robots,'' in \emph{IEEE/SICE International
  Symposium on System Integration (SII)}, 2019, pp. 805--810.

\bibitem{walker}
M.~Walker, H.~Hedayati, J.~Lee, and D.~Szafir, ``Communicating robot motion
  intent with augmented reality,'' in \emph{ACM/IEEE International Conference
  on Human-Robot Interaction}, 2018, p. 316–324.

\bibitem{dinh}
H.~Dinh, Q.~Yuan, I.~Vietcheslav, and G.~Seet, ``Augmented reality interface
  for taping robot,'' in \emph{2017 18th International Conference on Advanced
  Robotics (ICAR)}, 2017, pp. 275--280.

\bibitem{waymouth2021demonstrating}
B.~Waymouth, A.~Cosgun, R.~Newbury, T.~Tran, W.~P. Chan, T.~Drummond, and
  E.~Croft, ``Demonstrating cloth folding to robots: Design and evaluation of a
  2d and a 3d user interface,'' \emph{arXiv preprint arXiv:2104.02968}, 2021.

\bibitem{chan}
W.~P. Chan, C.~P. Quintero, M.~Pan, M.~Sakr, H.~Van~der Loos, and E.~Croft, ``A
  multimodal system using augmented reality, gestures, and tactile feedback for
  robot trajectory programming and execution,'' in \emph{Proceedings of the
  ICRA Workshop on Robotics in Virtual Reality, Brisbane, Australia}, 2018, pp.
  21--25.

\bibitem{chan2020}
W.~P. Chan, G.~Hanks, M.~Sakr, T.~Zuo, H.~Machiel Van~der Loos, and E.~Croft,
  ``An augmented reality human-robot physical collaboration interface design
  for shared, large-scale, labour-intensive manufacturing tasks,'' in
  \emph{2020 IEEE/RSJ International Conference on Intelligent Robots and
  Systems (IROS)}, 2020, pp. 11\,308--11\,313.

\bibitem{ARReview}
Z.~Makhataeva and H.~A. Varol, ``Augmented reality for robotics: a review,''
  \emph{Robotics}, vol.~9, no.~2, p.~21, 2020.

\bibitem{morris}
M.~Gu, A.~Cosgun, W.~Chan, T.~Drummond, and E.~Croft, ``Seeing thru walls:
  Visualizing mobile robots in augmented reality,'' in \emph{IEEE International
  Symposium on Robot and Human Interactive Communication (RO-MAN)}, 2021.

\bibitem{rhys}
R.~Newbury, A.~Cosgun, T.~Crowley-Davis, W.~P. Chan, T.~Drummond, and E.~Croft,
  ``Visualizing robot intent for object handovers with augmented reality,''
  \emph{arXiv preprint arXiv:2103.04055}, 2021.

\bibitem{moveit}
D.~T. Coleman, I.~A. Sucan, S.~Chitta, and N.~Correll, ``Reducing the barrier
  to entry of complex robotic software: a moveit! case study,'' \emph{JOSER -
  Journal of Software Engineering for Robotics}, vol.~5, pp. 3--16, 2014.

\end{thebibliography}

\end{document}